\documentclass[
]{ceurart}

\usepackage{graphicx}
\usepackage{framed}
\usepackage{changepage}

\usepackage{fancyvrb}
\VerbatimFootnotes

\begin{document}

\copyrightyear{2021}
\copyrightclause{Copyright for this paper by its authors.
  Use permitted under Creative Commons License Attribution 4.0
  International (CC BY 4.0).}

\conference{SemREC'21: Semantic Reasoning Evaluation Challenge, ISWC'21, Oct 24 -- 28, Albany, NY}

\title{The CaLiGraph Ontology as a Challenge for OWL~Reasoners}

\author[1]{Nicolas Heist}[%
orcid=0000-0002-4354-9138,
email=nico@informatik.uni-mannheim.de,
url=http://www.uni-mannheim.de/dws/people/researchers/phd-students/nicolas-heist/
]
\author[1]{Heiko Paulheim}[%
orcid=0000-0002-4354-9138,
email=heiko@informatik.uni-mannheim.de,
url=http://www.heikopaulheim.com/
]

\address[1]{Data and Web Science Group, University of Mannheim, Germany}

\begin{abstract}
CaLiGraph is a large-scale cross-domain knowledge graph generated from Wikipedia by exploiting the category system, list pages, and other list structures in Wikipedia, containing more than 15 million typed entities and around 10 million relation assertions. Other than knowledge graphs such as DBpedia and YAGO, whose ontologies are comparably simplistic, CaLiGraph also has a rich ontology, comprising more than 200,000 class restrictions. Those two properties -- a large A-box and a rich ontology -- make it an interesting challenge for benchmarking reasoners. In this paper, we show that a reasoning task which is particularly relevant for CaLiGraph, i.e., the materialization of \texttt{owl:hasValue} constraints into assertions between individuals and between individuals and literals, is insufficiently supported by available reasoning systems. We provide differently sized benchmark subsets of CaLiGraph, which can be used for performance analysis of reasoning systems.
\end{abstract}

\begin{keywords}
Knowledge Graph \sep Reasoning \sep Scalability \sep \texttt{owl:hasValue} Restrictions \sep Materialization
\end{keywords}

\maketitle

\section{Introduction}
In the recent decade, open cross-domain knowledge graphs have been recognized as an interesting ingredient to build intelligent systems. This gave rise to the development of various open knowledge graphs, such as DBpedia \cite{lehmann2015dbpedia}, YAGO \cite{suchanek2007yago}, and Wikidata \cite{vrandevcic2012wikidata}. While those knowledge graphs come with large-scale A-boxes comprising millions of entities, their T-boxes are usually not very expressive, defining mainly a class hierarchy and relations with domains and ranges, but not using complex class constructors \cite{heist2020knowledge}.

CaLiGraph is a comparatively new knowledge graph, which is constructed from categories, list pages, and other lists in Wikipedia. It uses DBpedia as a training set to derive interpretations of lists and categories \cite{heist2019uncovering}. This allows to extract more entities from list pages \cite{heist2020entity} and other listings \cite{heist2021information}, making the A-box comprise more than twice as many entities as DBpedia. At the same time, the definitions of the derived classes are also maintained in the T-box, making it more expressive than the ontologies of other knowledge graphs.

With those characteristics, CaLiGraph creates some interesting tasks to reasoners, since it requires the materialization of individual and literal assertions, and, at the same time, poses high scalability requirements when asking a reasoner to materialize 10 million A-box assertions. Since those requirements are hardly met by currently available reasoners, the materialization of the CaLiGraph A-box is carried out by custom code, not by running a reasoner, albeit being a standard materialization task which an OWL2 EL reasoner should be capable of performing.

In this paper, we report on experiments with three reasoning systems on the materialization of CaLiGraph. Furthermore, we introduce differently sized subsets of CaLiGraph, which allow for performing scalability experiments.

The rest of this paper is structured as follows. Section~\ref{sec:caligraph} gives a short introduction into CaLiGraph. In section~\ref{sec:experiments}, we describe a set of experiments we conducted to investigate the fitness of existing reasoners for processing CaLiGraph. We close with a short summary and an outlook on future work.

\begin{figure}[t]
    \centering
    \includegraphics[width=.96\textwidth]{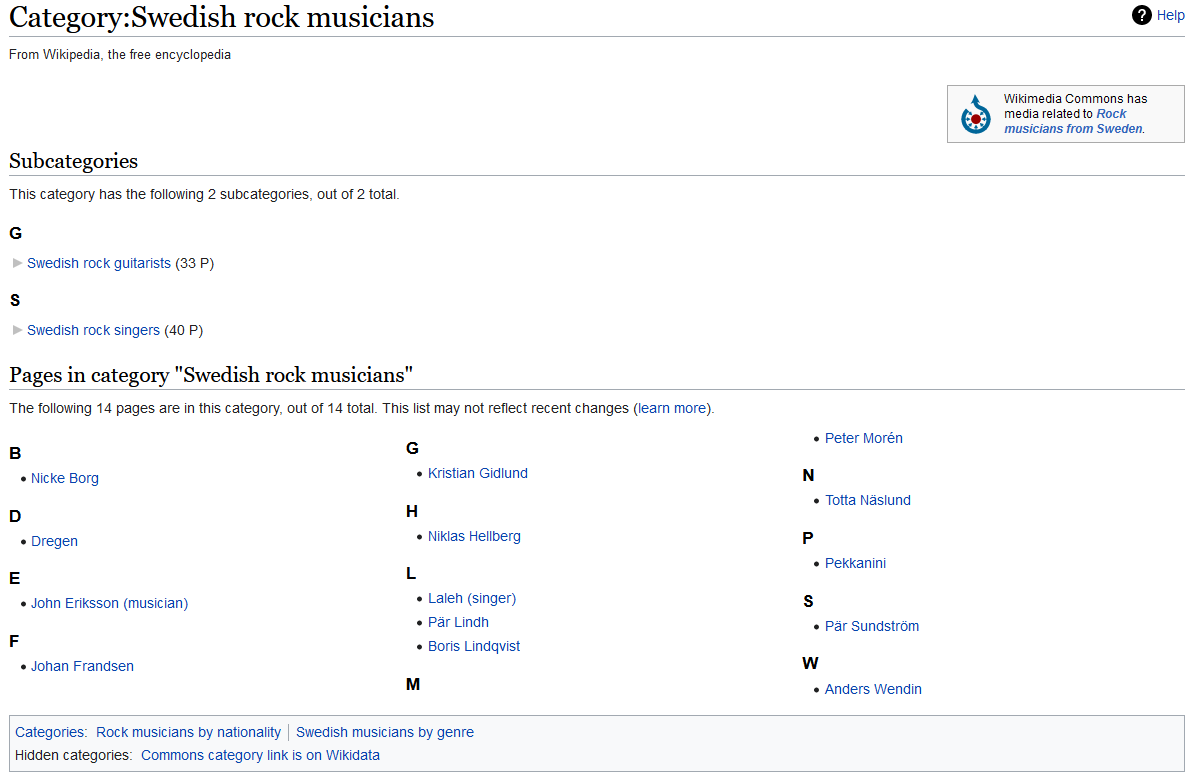}
    \caption{Example of a Category in Wikipedia}
    \label{fig:category_example}
\end{figure}

\section{The CaLiGraph Ontology}
\label{sec:caligraph}
As described above, CaLiGraph processes categories and list pages in Wikipedia, and derives classes and instances from those. For example, when processing the Wikipedia category \texttt{Swedish Rock Musicians},\footnote{\url{https://en.wikipedia.org/wiki/Category:Swedish_rock_musicians}} shown in Fig.~\ref{fig:category_example}, it will create a class \texttt{clgo:Swedish\_rock\_musician},\footnote{The following name space conventions are used throughout this paper:\newline\texttt{clgo=http://caligraph.org/ontology/}\newline\texttt{clgr=http://caligraph.org/resource/}} and heuristically assign the following subclass axioms to it:

\begin{Verbatim}[frame=single]
clgo:Swedish_rock_musician
  a owl:Class ;
  rdfs:subClassOf
    clgo:Rock_musician ,
    clgo:Swedish_musician ,
    [
      a owl:Restriction ;
      owl:onProperty clgo:birthPlace ;
      owl:hasValue clgr:Sweden
    ] ,
    [
      a owl:Restriction ;
      owl:onProperty clgo:genre ;
      owl:hasValue clgr:Rock_music
    ] ,
    [ 
      a owl:Restriction ;
      owl:onProperty clgo:occupation ;
      owl:hasValue clgr:Musician
    ] .

clgo:Swedish_rock_guitarist rdfs:subClassOf clgo:Swedish_rock_musician .
clgo:Swedish_rock_singer rdfs:subClassOf clgo:Swedish_rock_musician .
\end{Verbatim}

The subclass axioms are derived from the subcategory relations in Wikipedia, the restrictions are created using a
mix of statistic and textual heuristics \cite{heist2019uncovering,heist2020entity}.

As shown in this example, restrictions using \texttt{owl:hasValue} are quite frequently used in CaLiGraph. Overall, the current release has more than 200,000 of such restrictions, most of them using a resource as their object, but some also using literals (e.g., restrictions on birth years).

When populating the A-box, CaLiGraph creates entities for Wikipedia pages, as well as entities found on list pages. The only axioms created for those entities are the class memberships for the classes derived from the corresponding categories and list pages; the remaining axioms are materialized according to the class definitions. In the example above, given that the entity \texttt{Dennis Lyxzén} is found in the category \texttt{Swedish rock singers} (which is a subcategory of the above mentioned \texttt{Swedish rock musicians}, CaLiGraph will produce the following axiom:

\begin{verbatim}
clgr:Dennis_Lyxzén a clgo:Swedish_rock_singer .
\end{verbatim}
Given the T-box above, the derived axioms according to the definition above should include:
\begin{verbatim}
clgr:Dennis_Lyxzén a clgo:Swedish_rock_musician .
clgr:Dennis_Lyxzén a clgo:Rock_musician .
clgr:Dennis_Lyxzén a clgo:Swedish_musician .
clgr:Dennis_Lyxzén clgo:birthPlace clgr:Sweden .
clgr:Dennis_Lyxzén clgo:genre clgr:Rock_music .
clgr:Dennis_Lyxzén clgo:occupation clgr:Musician .
\end{verbatim}
The first three axioms are obtained by materializing the subclass relations, the latter three by materializing the restrictions based on \texttt{owl:hasValue}. Note that the example only depicts a subset of the information on the instance \texttt{clgr:Dennis\_Lyxzén} in CaLiGraph, which also contains information on his birthplace and birth year.\footnote{\url{http://caligraph.org/resource/Dennis_Lyxzén}}

While CaLiGraph uses a custom implementation to materialize those A-box axioms, this is a task which should, in principle, be possible to carry out for an OWL DL reasoner. Moreover, since restrictions with \texttt{owl:hasValue} are possible in OWL2 EL and OWL2 RL \cite{motik2009owl}, we would expect that reasoners supporting those fragments should be capable of carrying out that materialization as well \cite{antoniou2018survey}.

The characteristics of the most recent version 2.1.0 of CaLiGraph are shown in table~\ref{tab:caligraph_stats}. The table shows that a reasoner, provided with the 14 million individuals and their 50M direct type assertions, should in principle be able to derive 139M transitive type relations, as well as 10M relation and literal assertions.

\begin{table}[t]
    \centering
    \begin{tabular}{l|r}
         No. of classes & 1,061,598 \\
         No. of relations & 338 \\
         No. of subclass assertions & 1,711,270 \\
         No. of restrictions & \\
         ...with individuals & 214,248 \\
         ...with literals & 12,925 \\
         No. of classes with direct restrictions & 223,552 \\
         No. of classes with direct or inherited restrictions & 488,364 \\
         \hline
         No. of instances & 14,452,393 \\
         No. of direct type assertions & 50,169,052\\
         No. of transitive type assertions & 138,713,499\\
         No. of relation assertions & \\
         ...with individuals & 9,637,354 \\
         ...with literals & 969,022 \\
    \end{tabular}
    \caption{Statistics of the CaLiGraph ontology}
    \label{tab:caligraph_stats}
\end{table}

This kind of reasoning -- at least at that scale -- is rather underrepresented in current reasoning benchmarks, and often not evaluated at all. For example, in the ORE2015 benchmark, the task of materialization is defined as \emph{the computation of all instances for all named classes in the ontology}~\cite{parsia2017owl}. In other words: in that benchmark, the task is to infer all class assertions for an instance given all its literal and individual assertions, while for CaLiGraph, we require the reverse direction.

Among the 1,920 test cases in the ORE2015 benchmarks~\cite{parsia2017owl}, there are only 311 which use \texttt{owl:hasValue} constraints with an individual (max. 6,455 in a single file), and 279 which use \texttt{owl:hasValue} with a literal (max. 1,104 in a single file). The more recent OWL2Bench dataset~\cite{singh2020owl2bench}, proposed in 2020, contains exactly one \texttt{owl:hasValue} restriction, mainly for checking whether or not a reasoner supports that construct, but does not allow for evaluating that support at scale. In contrast, the CaLiGraph ontology makes much heavier use of those constructs, as shown above, and therefore imposes more challenging scalability requirements with respect to that reasoning task.

\section{Experiments}
\label{sec:experiments}
To validate how far reasoning systems actually meet those challenges, we conducted a series of experiments, both with CaLiGraph as a whole and with differently-sized subsets of CaLiGraph.

\subsection{Reasoning Systems}
In our experiments, we chose three commonly known reasoners: ELK \cite{kazakov2014elk}, HermiT \cite{glimm2014hermit}, and Pellet \cite{sirin2007pellet}. All of them claim to be able to process OWL2 EL ontologies. We used the latest non-commercial versions, i.e.:
\begin{itemize}
    \item ELK version 0.4.3
    \item Pellet version 2.3.6
    \item HermiT version 1.4.5.456
\end{itemize}
The reasoning systems were called through OWL API\footnote{\url{https://github.com/owlcs/owlapi}} using the respective connectors to the reasoners. As the reasoners rely on different versions of OWL API, we provide an individual dependency configuration for each of them in our implementation. Note that since OWL API does not support restrictions with explicit URIs and labels, we transformed them to blank nodes (as shown in Fig.~\ref{fig:sandbox_example}) before importing them with OWL API.

\subsection{Sandbox Example}
As a first sanity check, we extracted a simple sandbox example from CaLiGraph, which we coin \texttt{clg\_10} in the following. The sandbox example contains two classes with one \texttt{owl:hasValue} restriction each, one using an individual, one using a literal. For each of the two classes, there is an individual, so that an OWL2EL reasoner should be capable of inferring one individual and one literal assertion. The sandbox example is shown in Fig.~\ref{fig:sandbox_example}.

\begin{figure}[p]
\begin{framed}
\begin{scriptsize}
\begin{verbatim}
@prefix xsd: <http://www.w3.org/2001/XMLSchema#> .
@prefix rdfs: <http://www.w3.org/2000/01/rdf-schema#> .
@prefix owl: <http://www.w3.org/2002/07/owl#> .
@prefix clgo: <http://caligraph.org/ontology/> .
@prefix clgr: <http://caligraph.org/resource/> .

clgo:Agent a owl:Class;
  rdfs:subClassOf owl:Thing .

clgo:Organization a owl:Class;
  rdfs:subClassOf clgo:Agent .

clgo:Person a owl:Class;
  rdfs:subClassOf clgo:Agent;
  owl:disjointWith clgo:Organization .

clgo:International_organization a owl:Class;
  rdfs:subClassOf clgo:Organization .

clgo:Organization_based_in_Asia a owl:Class;
  rdfs:subClassOf clgo:Organization .

clgo:Organization_based_in_China a owl:Class;
  rdfs:subClassOf clgo:Organization_based_in_Asia .
  
clgo:headquarter a owl:ObjectProperty .

clgr:China a owl:NamedIndividual .

_:b1 a owl:Restriction;
  owl:onProperty clgo:headquarter;
  owl:hasValue clgr:China .

clgo:International_organization_based_in_China a owl:Class;
  rdfs:subClassOf clgo:Organization_based_in_China, clgo:International_organization, _:b1 .
  
clgr:International_Center_on_Small_Hydro_Power a owl:NamedIndividual,
                                                 clgo:International_organization_based_in_China .

clgo:Organization_disestablished_in_1939 a owl:Class;
  rdfs:subClassOf clgo:Organization .

clgo:activeYearsEndYear a owl:DatatypeProperty .

_:b2 a owl:Restriction;
  owl:onProperty clgo:activeYearsEndYear;
  owl:hasValue "1939"^^xsd:integer .

clgo:Military_unit_or_formation_disestablished_in_1939 a owl:Class;
  rdfs:subClassOf clgo:Organization_disestablished_in_1939, _:b2 .
  
clgr:46th_Mixed_Brigade a owl:NamedIndividual, clgo:Military_unit_or_formation_disestablished_in_1939 .
\end{verbatim}
\end{scriptsize}
\vspace{-1.0\baselineskip}
\end{framed}
\caption{Sandbox example for testing the reasoners' capabilities of processing restrictions (with individuals and literals) and disjointness axioms}
\label{fig:sandbox_example}
\end{figure}

From our three reasoners, only Pellet is capable of inferring both assertions. HermiT can infer the resource-valued assertion, but not the literal-valued one, while ELK does not infer any of the two. Table~\ref{tab:sandbox_results} summarizes the results of the sandbox experiment. Here, \emph{Disjointness materialization} refers to the addition of disjointness axioms to subclasses. In the example above, if \verb+clgo:Swedish_Rock_Musician+ was defined as a subclass of \verb+clgo:Person+, and \verb+clgo:Person+ is disjoint with \verb+clgo:Organization+, the reasoner would also infer that \verb+clgo:Swedish_Rock_Musician+ is disjoint with \verb+clgo:Organization+ (and its subclasses). However, we just inspected that out of curiosity, and removed disjointess axioms from the larger scale datasets.

\begin{table}[t]
    \centering
    \begin{tabular}{l|c|c|c}
         Reasoner &  ELK & HermiT & Pellet \\
         \hline
         Subclass materialization    & X & X & X \\
         Disjointness materialization &  & X & X \\
         Restrictions with individuals & & X & X \\
         Restrictions with literals & & & X \\
    \end{tabular}
    \caption{Reasoner capabilities as evaluated in the sandbox experiments}
    \label{tab:sandbox_results}
\end{table}

\begin{table}[t]
    \centering
    \begin{adjustwidth}{-0.25cm}{}
    \begin{tabular}{l|r|r|r|r||r|r|r}
    Dataset	&	Triples	&	Classes	&	Restrictions	&	Instances	& \multicolumn{3}{c}{Inferrable Assertions} \\
    	&		&		&		&		& Trans. Types & Individuals & Literals \\
    \hline
    \texttt{clg\_10}	&	35	&	9	&	2	&	3	&	10 & 1 & 1\\
    \texttt{clg\_10e2}	&	510	&	101	&	3	&	70	&	425 & 20 & 1\\
    \texttt{clg\_10e3}	&	43,120	&	1,000	&	79	&	18,072	&	4,500 & 514 & 5,706\\
    \texttt{clg\_10e4}	&	297,266	&	10,006	&	1,299	&	115,501	&	73,683 & 12,085 & 10,369\\
    \texttt{clg\_10e5}	&	4,641,400	&	99,923	&	12,147	&	1,968,282	&	42,820 & 187,929 & 7,120\\
    \texttt{clg\_full}	&	54,914,982	&	1,061,598	&	227,173	&	14,452,393	&	138,713,499 & 9,637,354 & 969,022\\
    \end{tabular}
    \end{adjustwidth}
    \caption{Overview of the different CaLiGraph subsets used for the evaluation}
    \label{tab:dataset_sizes}
\end{table}

We tried a couple of other reasoners, which were not capable of completely processing the Sandbox example, including
\begin{itemize}
    \item jcel \cite{mendez2012jcel} version 0.24.121
    \item Snorocket \cite{lawley2010fast} version 4.0.17
    \item Konclude \cite{steigmiller2014konclude} version 0.7.0
\end{itemize}
Those did either not produce any of the desired statements, or reported by themselves that they do not support inference based on \texttt{owl:hasValue} constructs. Thus, we decided to only keep ELK as a representative of those reasoners which cannot produce the inferences sought to compare its performance to those which can, and discarded the rest.

\subsection{Scalability Experiments}
To conduct experiments for scalability, we extracted differently sized subsets of CaLiGraph, each of which should have around $10^n$ classes for different values of $n$. The resulting datasets are depicted in table~\ref{tab:dataset_sizes}. The datasets were designed so that they create meaningful subtrees of the overall hierarchy in CaLiGraph. The corresponding SPARQL query is depicted in figure~\ref{fig:sparql_query_dataset_generation}. The different datasets are generated by varying the inner \texttt{SELECT} statement which retrieves the leaf classes. Datasets up to \texttt{clg\_10e4} are drawn from subclasses of \verb+clgo:Organization+, while \texttt{clg\_10e5} is drawn from the complete ontology. The relatively low number of transitive types of the latter dataset is explained by the fact that the average type depth of subclasses of \verb+clgo:Organization+ is much higher than the overall average type depth.

\begin{figure}[t]
\begin{framed}
\begin{scriptsize}
\begin{verbatim}
PREFIX owl: <http://www.w3.org/2002/07/owl#>
PREFIX rdfs: <http://www.w3.org/2000/01/rdf-schema#>
PREFIX clgo: <http://caligraph.org/ontology/>
CONSTRUCT {
    ?leaf ?leafPred ?leafObj .
    ?super ?superPred ?superObj .
    ?ind a owl:NamedIndividual, ?leaf .
    ?resPred ?resPredPred ?resPredObj .
    ?resObj ?resObjPred ?resObjObj .
} WHERE {
    # Collect set of leaf classes to bootstrap sample from
    # Sample subgraph can be changed by picking something else than clgo:Organization as basis
    # Use a specific LIMIT for N to restrict the number of leaf classes in the sample
    {
        SELECT ?leaf {
            ?leaf rdfs:subClassOf+ clgo:Organization .
            FILTER NOT EXISTS {?sub rdfs:subClassOf ?leaf}
        }
        LIMIT N
    }
    # Collect all relevant superclasses of the leaves
    ?leaf rdfs:subClassOf+ ?super ;
          ?leafPred ?leafObj .
    ?super ?superPred ?superObj .
    # Collect individuals for the leaves
    ?ind a ?leaf .
    # Add restrictions, if available
    OPTIONAL {
        ?leaf rdfs:subClassOf+ ?res .
        ?res a owl:Restriction ;
             owl:onProperty ?resPred ;
             owl:hasValue ?resObj .
        # Add information about the predicates and objects of restrictions to the sample
        ?resPred ?resPredPred ?resPredObj .
        OPTIONAL {
            ?resObj ?resObjPred ?resObjObj .
        }
    }
    # Discard disjointnesses as this would blow up the sample size
    FILTER (?superPred != owl:disjointWith)
    FILTER (?leafPred != owl:disjointWith)
}
\end{verbatim}
\end{scriptsize}
\vspace{-1.0\baselineskip}
\end{framed}
\caption{SPARQL query to generate datasets for scalability experiments}
\label{fig:sparql_query_dataset_generation}
\end{figure}

\begin{figure}[h]
    \centering
    \includegraphics[width=0.8\textwidth]{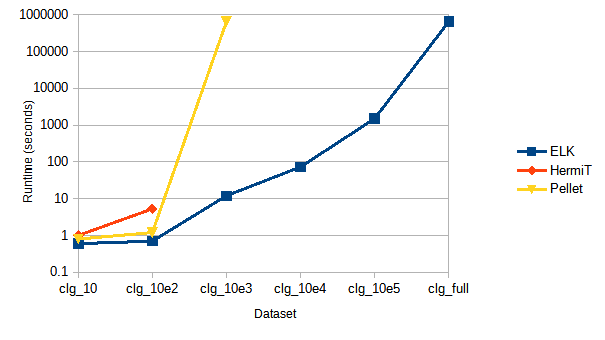}
    \caption{Scalability of the Reasoning Systems investigated}
    \label{fig:scalability}
\end{figure}

The reasoner performances are given in Fig.~\ref{fig:scalability}. We conducted experiments on a system with 32 Intel Xeon CPUs and 200GB RAM, and we set a timeout of 72h. Overall, ELK was the only system capable of processing all datasets (but not returning any of the inferrable individual and literal assertions, as discussed above). The processing of the largest dataset with ELK took about 3.5 hours.

HermiT and Pellet, on the other hand, could only process \texttt{clg\_10e2} and \texttt{clg\_10e3}, respectively, and timed out on the larger ones (with Pellet running out of memory on \texttt{clg\_full}). This shows that those reasoning systems are still three to four orders of magnitude away from the reasoning capabilities required for processing CaLiGraph.

\section{Conclusion and Outlook}
\label{sec:conclusion}
In this paper, we investigated three state of the art OWL2EL reasoners, i.e., ELK, HermiT, and Pellet. The task for those reasoners was to materialize the A-box of the CaLiGraph knowledge graph, particularly the creation of around 10M A-box assertions from \texttt{owl:hasValue} definitions. As discussed above, this kind of reasoning at scale is currently not evaluated in benchmarks such as ORE2015 and OWL2Bench.

The outcome of our experiments is that the reasoners evaluated are either (a) not capable of performing that inference at all, as in the case of ELK, or (b) not scalable enough to process knowledge graphs with 10k instances or more -- which is some orders of magnitude below the size of popular knowledge graphs \cite{heist2020knowledge}. 

With this work, we have also provided a benchmark suite of differently sized test sets.\footnote{The datasets and the code for running the experiments are available at \url{https://github.com/nheist/CaLiGraph-for-SemREC}.} This allows for running scalability experiments also with respect to materializing a large amount of individual and literal A-box assertions. Moreover, the benchmark is suitable for running evaluations with approximate reasoning systems, since recall and precision of the generated A-box assertions can be quantified. Therefore, the datasets are also suitable for measuring the trade-off between scalability and accuracy of approximate reasoning systems.

CaLiGraph is generated heuristically. Although the quality is high \cite{heist2021information}, it is not perfect. In the future, we want to further analyze the impact of wrong statements in the graph, as well as experiment with artificially adding noise to the different datasets.

While we have generated subsets of CaLiGraph using the SPARQL query in figure~\ref{fig:sparql_query_dataset_generation} and were therefore experimenting with manual limits to get the desired amount of classes for the sample, more sophisticated methods for generating subsets of knowledge graphs exist. For example, Melo and Paulheim \cite{melo2017synthesizing} propose a synthesis model for the generation of knowledge graph benchmark datasets which strives to preserve characteristics like class, relation and instance distributions. For future work, we consider to generate such more realistic benchmark datasets with the help of synthesis models.


\bibliography{references}

\end{document}